\documentclass[preprint,review,12pt]{elsarticle}

\usepackage{float}
\usepackage{amsmath}
\usepackage{graphicx}
\usepackage{caption}
\usepackage{subcaption}
\usepackage{tikz}
\usepackage{pifont}
\usetikzlibrary{shapes,arrows,positioning,fit,backgrounds}
\graphicspath{ {./} }

\usepackage[acronym]{glossaries}
\makeglossaries
\newacronym{llm}{LLM}{Large Language Model}

\usepackage{url}
\usepackage{xcolor}
\usepackage{hyperref}
\definecolor{newcolor}{rgb}{.8,.349,.1}

\journal{Pattern Recognition}

\newcommand{\red}[1]{\textcolor{red}{#1}}
\newcommand{\green}[1]{\textcolor{green!70!black}{#1}}
\newcommand{\blue}[1]{\textcolor{blue}{#1}}

\newcommand{\yescheckmark}{\textcolor{green!70!black}{\ding{51}}}
\newcommand{\xmark}{\textcolor{red}{\ding{55}}}

\begin{document}
    \begin{frontmatter}

        \title{GroundCap: A Visually Grounded Image Captioning Dataset}

        \author[inesc,ist]{Daniel A. P. Oliveira}
        \ead{daniel.oliveira@inesc-id.pt}

        \author[inesc]{Lourenço Teodoro}
        \ead{lourenco.teodoro@inesc-id.pt}

        \author[inesc,ist]{David Martins de Matos}
        \ead{david.matos@inesc-id.pt}

        \affiliation[inesc]{organization={INESC-ID Lisboa}, % Department and Organization
            addressline={R. Alves Redol 9},
            city={Lisboa},
            postcode={1000-029},
            country={Portugal}}

        \affiliation[ist]{organization={Instituto Superior Técnico, Universidade de Lisboa}, % Department and Organization
            addressline={Av. Rovisco Pais},
            city={Lisboa},
            postcode={1049-001},
            country={Portugal}
        }

        \begin{abstract}
            Current image captioning systems lack the ability to link descriptive text to specific visual elements, making their outputs difficult to verify.
            While recent approaches offer some grounding capabilities, they cannot track object identities across multiple references or ground both actions and objects simultaneously.
            We propose a novel ID-based grounding system that enables consistent object reference tracking and action-object linking.
            We present GroundCap, a dataset containing 52,016 images from 77 movies, with 344 human-annotated and 52,016 automatically generated captions.
            Each caption is grounded on detected objects (132 classes) and actions (51 classes) using a tag system that maintains object identity while linking actions to the corresponding objects.
            Our approach features persistent object IDs for reference tracking, explicit action-object linking, and
            the segmentation of background elements through K-means clustering.
            We propose gMETEOR, a metric combining caption quality with grounding accuracy, and establish baseline performance by fine-tuning Pixtral-12B and Qwen2.5-VL 7B on GroundCap.
            Human evaluation demonstrates our approach's effectiveness in producing verifiable descriptions with coherent object references.
        \end{abstract}

        \begin{graphicalabstract}
            \begin{figure*}[t]
                \centering
                \resizebox{\textwidth}{!}{
                    \input{figures/graphical-abstract}
                }
            \end{figure*}
        \end{graphicalabstract}

        \begin{highlights}
            \item Novel dataset GroundCap features 52,016 images with 52,350 grounded captions.

            \item Captions link actions or multiple objects, describing their interactions and roles.

            \item gMETEOR metric combines caption quality and grounding accuracy to evaluate captions.

            \item Fine-tuned Pixtral-12B and Qwen2.5VL-7B on GroundCap set strong captioning baselines.

            \item Human evaluation studies validate the quality of the grounded captions.
        \end{highlights}

        \begin{keyword}
            Grounded Captioning \sep Visual Grounding \sep Dataset \sep Object Detection \sep Visual-Linguistic Alignment

        \end{keyword}

    \end{frontmatter}

    \section{Introduction}\label{sec:introduction}
    One of the primary goals of combining computer vision and natural language processing is to enable machines to understand and communicate about visual scenes.
    This objective encompasses numerous tasks, including recognizing objects, describing their attributes and relationships, and providing contextually relevant descriptions of scenes~\cite{lin2014microsoft}.
    While significant progress has been made in image classification, object detection, and image captioning,
    a critical aspect of human visual communication remains under-explored: the ability to ground language to specific elements within an image.

    Consider a scenario where two people are discussing a crowded street scene.
    One might say, ``Look at that car.'' to which the other might respond, ``Which one?''.
    The first person would likely point to the specific car they're referring to while simultaneously describing it with more detail.
    This act of pointing while verbally describing and explicitly linking language to a visual element to resolve ambiguity is called grounding.
    Current models for image captioning and image understanding lack this intuitive ability to ground their descriptions on specific visual elements.
    This limitation can lead to ambiguous or misunderstood communications, particularly in complex visual scenarios
    with multiple objects and actions.

    We introduce GroundCap, a novel dataset and task formulation that addresses the challenge of visually grounded image captioning.
    GroundCap is designed to bridge this gap by providing a dataset of images with captions that are
    explicitly grounded to visual elements, mimicking the human ability to reference
    objects when describing a scene.

    GroundCap comprises 52,016 images extracted from 77 movies from the MovieNet dataset~\cite{huang2020MovieNetAH}.
    We consider two caption types: 52,016 automatically generated captions (42,016 for training and 10,000 for evaluation) and
    344 human-annotated captions (274 for training and 70 for evaluation).
    Each image is further annotated with object positions and classes from a set of 132 object classes.
    The captions are grounded using a novel tag format that links textual descriptions to multiple visual elements.
    These captions not only reference objects but also describe actions, and the actions themselves are grounded on one or more objects,
    providing for rich contextual descriptions.

    In contrast to works that focus solely on object grounding~\cite{peng2023kosmos, zhang2024groundhog},
    GroundCap introduces several key innovations.
    Our framework assigns unique identifiers to each object and unifies the grounding of objects, actions, and locations while maintaining
    consistent object identity throughout descriptions.
    This approach enables rich contextual descriptions that can reference multiple objects from different classes within
    a single text span and resolve complex references (e.g., linking pronouns like ``They'' to specific objects like ``person-1'' and ``person-2'').
    Our system can ground spans of text to specific portions of background elements, moving beyond treating them as single entities.

    Our main contributions can be summarized as follows:
    (1) a novel dataset of 52,016 images with grounded captions that maintain object identity across references;
    (2) a unified grounding framework that links objects, actions, and locations while handling background elements;
    (3) gMETEOR, a new metric that combines caption quality with grounding accuracy;
    (4) strong baseline models based on fine-tuned Pixtral-12B and Qwen2.5-VL 7B;
    (5) evaluation through automated metrics, human assessment, and ChatGPT-4o evaluation;
    The dataset and fine-tuned model are available on Hugging Face.

    The paper is organized as follows: Section~\ref{sec:related-work} reviews related work on image captioning,
    object detection, and grounded captioning.
    Section~\ref{sec:groundcap-dataset} details the creation of GroundCap.
    Section~\ref{sec:gmeteor} introduces gMETEOR, a novel metric designed to assess the quality of grounded captions.
    Section~\ref{sec:baseline-model} presents baseline models for GroundCap.
    Section~\ref{sec:human-evaluation} outlines the results of human evaluation of GroundCap, baseline model and human-refined captions.
    Section~\ref{sec:limitations} outlines current limitations.
    Section~\ref{sec:conclusions} presents the conclusions and future work.

    \section{Related Work}\label{sec:related-work}
    In the context of grounded captioning, several tasks and their associated datasets are particularly relevant, as discussed in
    the following sections.

    The development of object recognition began with datasets like Caltech 101~\cite{fei2004learning},
    which established foundations for categorizing visual objects.
    A significant leap forward was made with ImageNet~\cite{deng2009imagenet}, containing millions of images across thousands of categories, enabling breakthroughs in deep learning for visual recognition~\cite{krizhevsky2012imagenet}.
    Object detection datasets build upon classification by requiring localization of objects.
    Notable examples include PASCAL VOC~\cite{everingham2010pascal} and MS COCO~\cite{lin2014microsoft}, which provide bounding
    box annotations for multiple object categories in natural images.
    The RefCOCO datasets~\cite{refitgame2014, yu2016modeling} further extended object localization
    by introducing natural language descriptions to guide object selection.
    Recent work~\cite{tong2023rethinking} has focused on improving annotation quality of existing benchmarks.

    Image captioning datasets combine visual and textual information, requiring models to generate descriptive sentences for images.
    Flickr30k~\cite{young2014image} and MS COCO Captions~\cite{lin2014microsoft} are widely used benchmarks in this domain.
    These datasets, however, typically provide only general descriptions without explicit grounding to specific image regions or objects.
    Visual Genome~\cite{krishna2017visual} takes a step towards grounded captioning by providing several region-specific
    captions for each image, where each caption describes a particular area defined by a single bounding box.
    While this approach offers more localized descriptions, it differs from our task in GroundCap.
    DenseCap~\cite{johnson2016densecap} introduced dense captioning for localizing and describing salient regions in images.
    Recent advances in Multimodal Large Language Models, such as Flamingo~\cite{alayrac2022flamingo},
    LLaVA~\cite{liu2023visual}, Pixtral-12B~\cite{pixtral12B} and Qwen2.5-VL~\cite{qwen25vl} have significantly improved image captioning capabilities.
    While some of these models, like Qwen-VL, provide visual grounding capabilities, they lack the ability to maintain
    consistent object identities across multiple references within a caption.

    Several works have explored aspects of grounded captioning, focusing on linking textual descriptions to visual elements.
    Visual Genome~\cite{krishna2017visual}, provides multiple region descriptions per image, each associated with a single bounding box.
    This approach limits the ability to create dense descriptions where words can be simultaneously grounded to multiple objects and actions across regions if the frame.
    The region-specific approach also constrains the ability to capture scene descriptions that integrate both static elements and dynamic interactions within a caption.
    Recent work has explored different approaches to visual grounding using \glspl{llm}.
    GROUNDHOG~\cite{zhang2024groundhog} uses pixel-level segmentation with unified \texttt{<GRD>} tags,
    Groma~\cite{ma2024groma} employs region tokens for arbitrary region grounding,
    and KOSMOS-2~\cite{peng2023kosmos} represents coordinates as discrete location tokens in a Markdown-like format.
    Florence-2~\cite{xiao2023Florence2AA} performs grounding through a sequence-to-sequence architecture that quantizes
    coordinates into 1,000 bins and represents regions in various formats (box, quad box, or polygon) depending on the task
    requirements.
    Panoptic Narrative Grounding (PNG)~\cite{gonzalez2021panoptic} grounds descriptions to panoptic segmentation regions
    using mouse trace annotations and WordNet-based~\cite{wordnet1994} semantic matching.
    \cite{cai2024top} proposes a one-stage weakly-supervised approach that processes raw RGB images and computes visual language attention maps.

    Several large-scale grounded datasets have been developed to support these approaches.
    GLaMM~\cite{rasheed2024glamm} introduced GranD-f, a dataset of 214K image-grounded text pairs.
    Groma~\cite{ma2024groma} curated Groma Instruct with 30K visually grounded conversations.
    KOSMOS-2~\cite{peng2023kosmos} was trained on GrIT, a dataset of 91M images with grounded image-text pairs, though only a small subset is publicly available.
    Florence-2 was trained on FLD-5B~\cite{xiao2023Florence2AA}, containing 126M images with 5.4B annotations, but the dataset remains proprietary.
    GROUNDHOG utilizes the M3G2 dataset with 2.5M images, which is also not publicly available.
    The availability of training data remains a significant factor in the reproducibility and advancement of grounded captioning research.
    Despite their promising performance, these models lack the ability to track object identities across multiple references
    or ground both actions and objects simultaneously.

    To position GroundCap among the existing datasets, Table~\ref{tab:dataset-comparison} presents a systematic comparison
    across key dimensions including scale, grounding capabilities, annotation types, and evaluation metrics.

    \begin{table*}[t]
        \centering
        \caption{Comparison of GroundCap with related grounded captioning and visual grounding datasets.
        Avg Words refers to the average number of words per caption.
        Objects indicates whether the dataset includes object annotations.
        Actions indicates whether the dataset includes action annotations.
        Object Re-ID indicates whether the dataset can consistently track and ground all instances of the same object across multiple references within captions.
        Public refers to whether the full dataset is publicly available.}
        \label{tab:dataset-comparison}
        \resizebox{\textwidth}{!}{
            \begin{tabular}{l|c|c|c|c|c|c}
                \hline
                \textbf{Dataset}                               & \textbf{\# Images} & \textbf{Objects} & \textbf{Actions} & \textbf{Object Re-ID} & \textbf{Avg Words} & \textbf{Public} \\
                \hline
                \hline
                Flickr30k Entities~\cite{plummer2015flickr30k} & 31,783             & \yescheckmark    & \xmark           & \xmark                & -                  & \yescheckmark \\
                Visual Genome~\cite{krishna2017visual}         & 108,249            & \yescheckmark    & \xmark           & \xmark                & 5                  & \yescheckmark   \\
                DenseCap~\cite{johnson2016densecap}            & 94,000             & \yescheckmark    & \xmark           & \xmark                & -                  & \yescheckmark   \\
                RefCOCO+~\cite{yu2016modeling}                 & 19,992             & \yescheckmark    & \xmark           & \xmark                & 3.5                & \yescheckmark   \\
                RefCOCOg~\cite{mao2016generation}              & 25,799             & \yescheckmark    & \xmark           & \xmark                & 8.4                & \yescheckmark   \\
                KOSMOS-2 (GrIT)~\cite{peng2023kosmos}          & 91,000,000         & \yescheckmark    & \xmark           & \xmark                & -                  & \xmark          \\
                GROUNDHOG (M3G2)~\cite{zhang2024groundhog}     & 2,500,000          & \yescheckmark    & \xmark           & \xmark                & -                  & \xmark \\
                GLaMM (GranD-f)~\cite{rasheed2024glamm}        & 214,000            & \yescheckmark    & \xmark           & \xmark                & -                  & \yescheckmark   \\
                Groma Instruct~\cite{ma2024groma}              & 30,000             & \yescheckmark    & \xmark           & \xmark                & -                  & \yescheckmark   \\
                Florence-2 (FLD-5B)~\cite{xiao2023Florence2AA} & 126,000,000        & \yescheckmark    & \xmark           & \xmark                & -                  & \xmark          \\
                \hline
                \textbf{GroundCap}                             & \textbf{52,016}    & \yescheckmark    & \yescheckmark    & \yescheckmark         & \textbf{128}       & \yescheckmark \\
                \hline
            \end{tabular}}
    \end{table*}

    \section{The GroundCap Dataset}\label{sec:groundcap-dataset}
    This section details the creation of GroundCap~\footnote{The dataset is available for academic research under the same license as the underlying MovieNet dataset at \url{https://huggingface.co/datasets/daniel3303/GroundCap}}.
    We first describe our frame selection process from MovieNet, followed by the object detection approach that handles both distinct objects and background elements.
    We then explain the caption generation pipeline that combines automated generation with human refinement.

    While current approaches focus primarily on object and region grounding, our semantically differentiated tag system
    (\texttt{<gda>} for actions, \texttt{<gdl>} for locations, \texttt{<gdo>} for objects) combined with object IDs
    allows us to ground actions to specific objects (e.g., linking the action ``running'' directly to ``person-1''),
    enabling a richer understanding of scene dynamics and interactions.

\begin{figure}[htpb]
        \centering
        \includegraphics[width=\columnwidth]{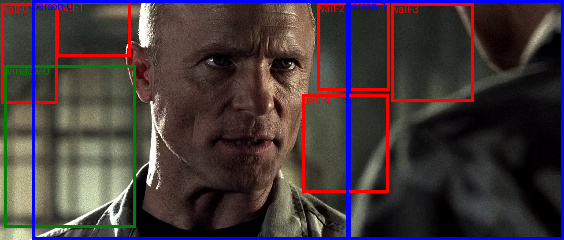}
        \caption[GroundCap Example]{Example of image from GroundCap. The grounded caption automatically generated by our baseline is in the main text due to space constraints.
        We can see that the model in able to ground actions, objects and locations on the corresponding \texttt{<gda>},
            \texttt{<gdo>} and \texttt{<gdl>} tags.
        }
        \label{fig:example}
    \end{figure}

    Figure~\ref{fig:example} shows an example of a grounded caption from our dataset.
    The grounding tags (\texttt{<gdo>}, \texttt{<gda>}, \texttt{<gdl>}) link text spans to specific objects, actions, and locations in the image,
    while maintaining object identity through unique IDs (e.g., person-0, person-1).
    This enables explicit tracking of objects across multiple references and links actions to their performing objects.

    {\sloppy
    ``In this dimly lit room,
        \texttt{<gdo class="person" \blue{person-0}>a bald man</gdo>}
        \texttt{<gda class="frown" \blue{person-0}>frowns</gda>} with a serious expression. He is positioned near
        \texttt{<gdl class="wall" \red{wall-0 wall-1 wall-2}>the walls</gdl>} of the room, which are adorned with
    \texttt{<gdo class="window" \green{window-0}>windows</gdo>} that let in a small amount of light.
    \texttt{<gdo class="person" \blue{person-1}>Another individual</gdo>} is partially visible on the right side of the image.
    The overall atmosphere is one of quiet intensity, with the
    \texttt{<gdo class="person" \blue{person-0}>man's</gdo>} expression suggesting a moment of contemplation or concern.''
    \par}

    \subsection{Movie, Scene, and Frame Selection from MovieNet}\label{subsec:frame-selection}
    The visual content for GroundCap is sourced from MovieNet~\cite{huang2020MovieNetAH} through a systematic selection process.
    %\subsubsection{Movie and Scene Selection}\label{subsubsec:movie-and-scene-selection}
    We start by selecting movies that have rich metadata, i.e., movies that include non-empty
    fields for plot, storyline, synopsis, and overview.
    For each movie, we apply several criteria to identify suitable scenes: they must contain both action and place tags, ensuring a rich semantic context for captioning.
    Additionally, scenes must include shots where characters are present, as determined by the cast annotations.
    This requirement ensures that the dataset captures human activities and interactions, avoiding purely static scenes.
    To mitigate potential noise from opening and closing credits, we exclude the first and last scenes of each movie from our selection.
    This selection process resulted in a dataset comprising 77 movies and 6,315 scenes, with an average of 82 scenes per movie.

%    \subsubsection{Video Frame Extraction}\label{subsubsec:video-frame-extraction}
    From the set of selected scenes, we use a systematic method to extract the frames, keeping only the shots that contain characters.
    From this set, we select up to 17, distributed evenly throughout the scene's duration,
    so that the total number of frames is approximately 50,000.
    From each identified shot, we extract the middle frame.
    When a scene contains fewer than 17 shots, we include all available shots, extracting the middle frame from each.
    This allows us to maintain a good representation of shorter scenes without oversampling from them.
    The frame extraction process yielded a total of 52,478 frames.
    From this set, we removed 461 frames that either contained no objects, were completely black, or were blurry.
    This resulted in a final dataset of 52,016 frames, with an average of 8.2 frames per scene and 676 frames per movie.

    \subsection{Object Detection}\label{subsec:object-detection}

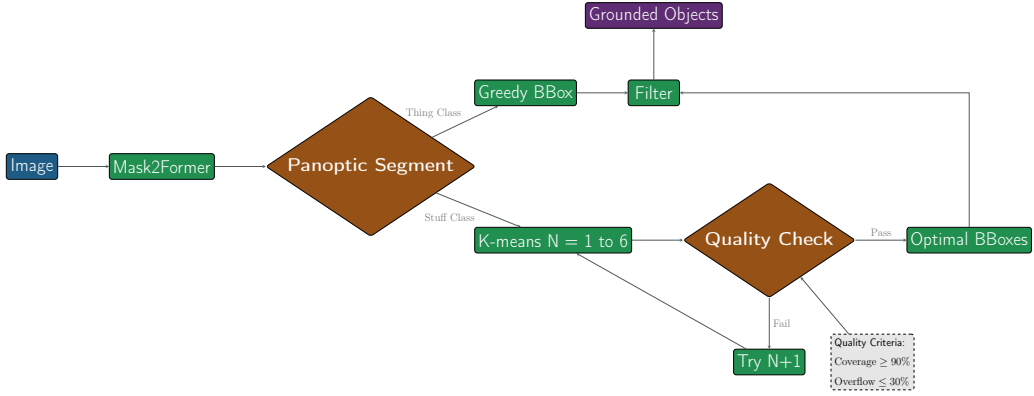
\begin{figure*}[t]
        \centering
        \resizebox{\textwidth}{!}{
            \begin{tikzpicture}[
    node distance=2cm,
    % Modern flat style definitions
    basenode/.style={
        draw,
        rectangle,
        rounded corners=3pt,
        minimum height=1cm,
        minimum width=2cm,
        text=white,
        font=\sffamily\LARGE
    },
    inputnode/.style={
        basenode,
        fill={rgb:red,41;green,128;blue,185}  % Modern blue
    },
    processnode/.style={
        basenode,
        fill={rgb:red,46;green,204;blue,113}  % Modern green
    },
    decisionnode/.style={
        basenode,
        diamond,
        aspect=1.5,
        fill={rgb:red,230;green,126;blue,34},  % Modern orange
        minimum height=1.5cm,
        align=center
    },
    outputnode/.style={
        basenode,
        fill={rgb:red,142;green,68;blue,173}  % Modern purple
    },
    arrow/.style={
        ->,
        >=stealth,
        thick,
        color={rgb:red,149;green,165;blue,166}  % Modern gray
    }
]
    % Input node
    \node[inputnode] (input_image) {Image};

    % Process nodes
    \node[processnode, right=of input_image] (mask2former) {Mask2Former};
    \node[decisionnode, right=of mask2former] (panoptic_segment) {\textbf{Panoptic Segment}};

    % Upper branch
    \node[processnode, above right=1cm and 2cm of panoptic_segment] (direct_box) {Greedy BBox};

    % Lower branch
    \node[processnode, below right=1cm and 2cm of panoptic_segment] (kmeans) {K-means N = 1 to 6};
    \node[decisionnode, right=2cm of kmeans] (quality_check) {\textbf{Quality Check}};
    \node[processnode, below=of quality_check] (increment_n) {Try N+1};
    \node[processnode, right=2cm of quality_check] (optimal_boxes) {Optimal BBoxes};

    % Final nodes
    \node[processnode, right=2cm of direct_box] (filter) {Filter};
    \node[outputnode, above=2cm of filter] (grounded_objects) {Grounded Objects};

    % Quality criteria box with modern styling
    \node[draw, dashed, rounded corners=3pt, fill=black!10,
          align=left, right=1cm of increment_n, text=black, font=\sffamily] (criteria)
          {Quality Criteria: \\$\textrm{Coverage} \geq 90\%$ \\ $\textrm{Overflow} \leq 30\%$};

    % Connect nodes with modern arrows
    \draw[arrow] (input_image) -- (mask2former);
    \draw[arrow] (mask2former) -- (panoptic_segment);
    \draw[arrow] (panoptic_segment) -- node[above left, text=gray] {Thing Class} (direct_box);
    \draw[arrow] (panoptic_segment) -- node[below left, text=gray] {Stuff Class} (kmeans);
    \draw[arrow] (kmeans) -- (quality_check);
    \draw[arrow] (quality_check) -- node[right, text=gray] {Fail} (increment_n);
    \draw[arrow] (increment_n) -- (kmeans);
    \draw[arrow] (quality_check) -- node[above, text=gray] {Pass} (optimal_boxes);
    \draw[arrow] (direct_box) -- (filter);
    \draw[arrow] (optimal_boxes) |- (filter);
    \draw[arrow] (filter) -- (grounded_objects);
    \draw[arrow] (criteria) -- (quality_check);
\end{tikzpicture}
        }
        \caption{Object detection pipeline overview.
        The process begins with Mask2Former segmentation and splits into separate paths for ``thing'' and ``stuff'' classes.
        ``Thing'' classes undergo direct bounding box generation. ``Stuff'' classes undergo an iterative K-means clustering process. Both paths converge in a filtering and sorting post-processing stage.
        }
        \label{fig:object-detection-diagram}
    \end{figure*}

    GroundCap uses an object detection approach based on panoptic segmentation using
    Mask2Former~\cite{cheng2021mask2former} with a Swin-Large backbone~\cite{liu2021swin} pre-trained on
    the MS COCO Panoptic dataset~\cite{lin2014microsoft}, providing a good balance between model size and SOTA accuracy.
    This method enables simultaneous detection and segmentation of both distinct objects (``thing'' classes)
    and background elements (``stuff'' classes).
    Fig.~\ref{fig:object-detection-diagram} shows the object detection pipeline.

    \subsubsection{Detection Process}\label{subsubsec:detection-process}
    The object detection process begins with images being passed through the Mask2Former model, generating panoptic segmentation outputs.
    From these outputs, we extract detections using different approaches for ``thing'' and ``stuff'' classes.
    For ``thing'' classes, representing distinct objects, bounding boxes are directly computed using a greedy approach
    that sets the minimum bounding box that includes the entire segmented area.
    For ``stuff'' classes, typically representing background elements like sky or buildings, we use a novel
    K-means clustering approach to generate multiple bounding boxes when appropriate, allowing for more granular detection
    of background elements.
    This approach is required because, in most cases, a greedy approach would produce a single bounding box containing most of the image.

    \subsubsection{Bounding Box Generation for ``Stuff'' Classes}\label{subsubsec:bounding-box-generation-for-stuff-classes}
    Our approach to generate bounding boxes for ``stuff'' classes aims to provide a detailed representation of background
    elements while balancing with the lowest number of bounding boxes possible.
    The process begins by extracting pixel coordinates of the segmented area.
    K-means clustering is then applied to these coordinates, with K ranging from 1 to 6.

    For each clustering result, we calculate the 5th and 95th percentiles of the $x$ and $y$ coordinates within each cluster.
    These percentiles form the basis of our initial bounding boxes, excluding the 5\% most extreme points on each end.
    This approach strikes a balance between coverage and compactness, ensuring that the initial boxes are representative
    of the cluster's main body without being influenced by scattered outlier points.

    After generating these initial boxes, we perform an adjustment step to eliminate overlaps between boxes of the same class.
    Our adjustment algorithm iteratively identifies overlapping pairs of boxes and modifies their boundaries.
    When adjusting for overlaps, we prioritize minimal distortion by modifying the overlapping edges proportionally to
    the size of each box.
    We prefer adjusting the lowest overlapping axis with respect to the length of the boxes in the corresponding axis.

    For instance, we compare the horizontal overlap relative to the total width of the overlapping boxes against
    the vertical overlap relative to their total height.
    If the proportional horizontal overlap is smaller, we adjust the left and right edges; otherwise, we adjust the top and bottom edges, thus preserving the aspect ratio of the boxes as much as possible.
    The adjustment for each box is calculated based on its size relative to the total size of both overlapping
    boxes in the chosen dimension (i.e., larger boxes undergo larger adjustments).
    The process results in a set of non-overlapping boxes that adhere to the underlying clusters,
    balancing the need for coverage of the segmented area with minimal overflow into non-segmented regions
    and minimizing the number of bounding boxes.
    The final set of boxes is evaluated based on two primary criteria.
    
    \textbf{Coverage}: The proportion of the segmented area that is enclosed by the bounding boxes. We aim for a minimum coverage of 90\% to ensure that the majority of the ``stuff'' region is represented.
    \textbf {Overflow}: The proportion of the area within the bounding boxes that does not belong to the segmented region. We aim for a maximum allowed overflow of 30\% into non-segmented areas.

    The evaluation function balances these criteria, seeking to maximize coverage while minimizing overflow.
    This process is repeated 10 times for each number of clusters until a valid configuration is found.
    A configuration is considered valid when the coverage exceeds 90\% and the overflow is below 30\% or
    when the maximum number of clusters is reached.
    If multiple valid configurations are found for the same number of clusters, the best one is selected based on
    the configuration score calculated as the difference between coverage and overflow.
    This allows us to represent large or disconnected background elements with multiple focused bounding boxes.

    Using multiple boxes allows to capture the shape and extent of ``stuff'' regions more accurately
    than a single bounding box could, providing a more detailed representation of the background elements in each image.
    The flexibility of this method enables it to adapt to various shapes and distributions of ``stuff'' classes,
    from large continuous areas like sky or road to more fragmented or complex shapes like groups of trees or building facades.

    \subsubsection{Post-processing}\label{subsubsec:post-processing}
    After generating all detections we sort them based on their confidence scores and only the top 40 most likely
    detections to avoid cluttering.
    Based on empirical evidence all detected objects with a confidence score below 0.7 are also discarded to
    ensure that only high-quality detections are retained.

    After filtering, the detected objects are sorted to establish a consistent reading order in the image,
    similar to how text is read.
    To achieve this, we implement a multi-level sorting strategy that first divides the image into three horizontal bands
    (by splitting the normalized y-coordinate in the bands 0-0.33, 0.33-0.66, and 0.66-1),
    and then sorts objects by their position.
    Specifically, objects are first sorted by their band position (top to bottom), then by their horizontal position within each band (left to right).

    We assign unique IDs to each detected object, following their spatial sorting order: e.g., person-0 will always appear before person-1 in the image, either in a higher band (closer to the top) or in the same band but further to the left.
    The consistent ordering ensures that object references in captions are predictable and spatially meaningful {--} when referring to ``the person in the top left corner'', we can be confident it corresponds to person-0.

    \subsection{Grounded Caption Generation}\label{subsec:grounded-caption-generation}

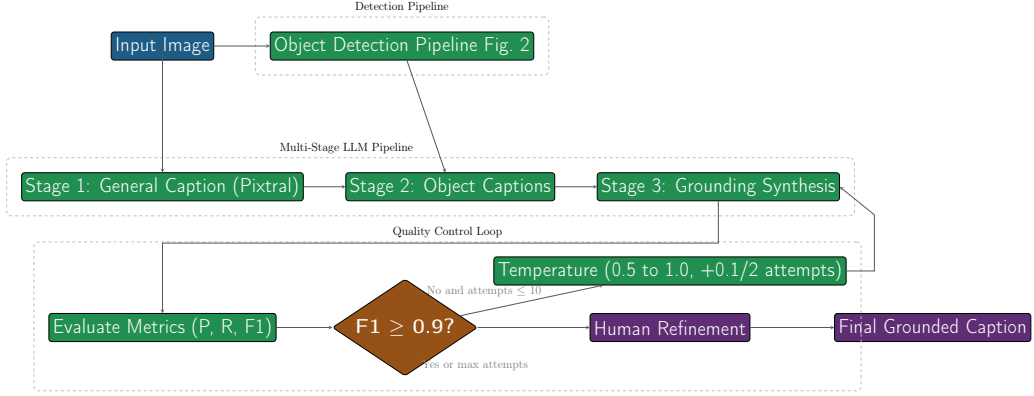
\begin{figure*}[t]
        \centering
        \resizebox{\textwidth}{!}{
            \begin{tikzpicture}[
    node distance=2cm,
% Modern flat style definitions
    basenode/.style={
        draw,
        rectangle,
        rounded corners=3pt,
        minimum height=1cm,
        minimum width=2cm,
        text=white,
        font=\sffamily\LARGE
    },
    inputnode/.style={
        basenode,
        fill={rgb:red,41;green,128;blue,185}  % Modern blue
    },
    processnode/.style={
        basenode,
        fill={rgb:red,46;green,204;blue,113}  % Modern green
    },
    decisionnode/.style={
        basenode,
        diamond,
        aspect=1.5,
        fill={rgb:red,230;green,126;blue,34},  % Modern orange
        minimum height=1.5cm,
        align=center
    },
    outputnode/.style={
        basenode,
        fill={rgb:red,142;green,68;blue,173}  % Modern purple
    },
    arrow/.style={
        ->,
        >=stealth,
        thick,
        color={rgb:red,149;green,165;blue,166}  % Modern gray
    },
    group/.style={
        draw=gray!50,
        dashed,
        rounded corners=5pt,
        inner sep=15pt,
        fit=#1
    }
]
    % Detection group nodes
    \node[inputnode] (input) {Input Image};
    \node[processnode, right=2cm of input] (detect) {Object Detection Pipeline Fig.~\ref{fig:object-detection-diagram}};

    % LLM Pipeline nodes
    \node[processnode, below=4cm of input] (stage1) {Stage 1: General Caption (Pixtral)};
    \node[processnode, right=1.5cm of stage1] (stage2) {Stage 2: Object Captions};
    \node[processnode, right=1.5cm of stage2] (stage3) {Stage 3: Grounding Synthesis};

    % Quality Control nodes
    \node[processnode, below=4cm of stage1] (eval) {Evaluate Metrics (P, R, F1)};
    \node[decisionnode, right=2cm of eval] (check) {\textbf{F1 $\geq$ 0.9?}};

    % Human and Output nodes
    \node[outputnode, right=4cm of check] (human) {Human Refinement};
    \node[outputnode, right=3cm of human] (output) {Final Grounded Caption};

    % Temperature node
    \node[processnode, above=1cm of human] (temp) {Temperature (0.5 to 1.0, +0.1/2 attempts)};

    % Groups with modern styling
    \node[group={(detect)}, label=above:Detection Pipeline] {};  % Updated to only include detect
    \node[group={(stage1) (stage2) (stage3)}, label={[xshift=-3cm]above:Multi-Stage LLM Pipeline}] {};  % Shifted label left
    \node[group={(eval) (check) (temp)}, label=above:Quality Control Loop] {};

    % Connections with modern arrows
    \draw[arrow] (input) -- (detect);
    \draw[arrow] (input) -| (stage1);
    \draw[arrow] (detect) -- (stage2);
    \draw[arrow] (stage1) -- (stage2);
    \draw[arrow] (stage2) -- (stage3);
    \draw[arrow] (stage3.south) -- ++(0,-1.5) -| (eval);
    \draw[arrow] (eval) -- (check);
    \draw[arrow] (check) -- node[above left=0cm and -0.5cm, text=gray] {No and attempts $\le$ 10} (temp);
    \draw[arrow] (check) -- node[below left=1cm and 0.05cm, text=gray] {Yes or max attempts} (human);
    \draw[arrow] (temp.east) -| ++(1,2) -- (stage3.east);
    \draw[arrow] (human) -- (output);
\end{tikzpicture}
        }
        \caption{Grounded caption generation pipeline overview.
        The process begins with Mask2Former object detection, followed by a multi-stage \gls{llm} pipeline using Pixtral for general scene description and object-specific captions.
        The quality control loop evaluates captions using precision, recall, and F1 metrics, with iterative refinement through temperature adjustment (0.5 to 1.0) for up to 10 attempts or until F1 $\geq$ 0.9.
        The pipeline concludes with human refinement to ensure caption quality and grounding accuracy.}
        \label{fig:ground-cap-creation-pipeline}
    \end{figure*}

    The grounded caption generation process (fig.~\ref{fig:ground-cap-creation-pipeline}) uses a multi-stage approach that combines \glspl{llm} with iterative refinement to ensure high-quality, accurately grounded captions.
    Following recent advances in vision-language models~\cite{liu2023visual,li2024llava},
    we implement a 3-stage pipeline: initial caption generation, object-specific captioning, and grounding synthesis.

    The process begins by generating a general caption for the main image using Pixtral~\cite{pixtral12B}.
    This initial caption provides a high-level description of the scene without explicit grounding information.
    The model receives the image along with a system prompt explaining the captioning task and a set of few-shot examples demonstrating proper caption format and style, following the few-shot learning paradigm established in recent work~\cite{liu2023visual}.
    We then generate focused captions for each detected object.
    These object-specific captions are created by providing the model with cropped regions corresponding
    to the bounding box of each object.
    In the final stage, the \gls{llm} synthesizes a grounded caption by combining the main caption, object-specific captions, and detection information, such as object metadata, including class, ID,
    and coordinates of the bounding box.
    The model generates a grounded caption incorporating our grounding tag system, using three distinct tags: \texttt{<gdo>} for objects, \texttt{<gda>} for actions, and \texttt{<gdl>} for locations.

    Quality control is implemented through an evaluation system based on precision ($P$), recall ($R$), and F1 score metrics.
    These values follow standard information retrieval definitions: $P=\text{TP}/(\text{TP}+\text{FP})$; $R=\text{TP}/(\text{TP}+\text{FN})$; and F1 as the harmonic mean of precision and recall. TP (true positives) represents correctly grounded objects, FP (false positives) are objects mentioned
    in the caption but not present in detections, and FN (false negatives) are detected objects omitted from the caption.

    Our iterative refinement process has a minimum F1 threshold of 0.9.
    When a caption fails to meet the threshold, error analysis generates feedback for the model.
    For adjusting $P$, we identify incorrectly referenced objects and prompt their removal or correction.
    For adjusting $R$, we highlight detected objects missing from the caption.
    We also monitor tag syntax, providing feedback on malformed grounding tags, such as missing class attributes
    or invalid object IDs.
    The refinement process continues for up to ten iterations, with the system retaining the highest-scoring
    caption throughout.
    To encourage exploration of different caption formulations~\cite{holtzman2019curious},
    the temperature parameter is set to 0.5 and is increased by 0.1 every two attempts until it reaches a maximum value of 1.0.
    The process ends when either a caption meets the F1 threshold or the maximum number of attempts is reached.
    The F1 score evaluates how well the generated caption grounds its descriptions to the detected objects, ensuring that
    the captions are linked to visual elements in the scene.
    However, while the F1 score measures whether objects are referenced in the caption, it does not verify if the grounding tags encompass the semantically correct spans of text, i.e., a caption might achieve a perfect F1 score while incorrectly grounding objects to the wrong spans of text.

    \subsection{Human Refinement}\label{subsec:human-annotation}

    We implemented a human refinement stage to ensure caption quality and accuracy.
    Rather than having annotators write captions from scratch, however, human annotators review and improve machine-generated captions.
    Annotators were provided detailed guidelines and instructed to preserve the original caption's structure when possible, making targeted improvements rather than complete rewrites unless necessary.
    The refinement-based approach reduces annotation time, while providing a structural template that helps maintain uniform style across the dataset.
    The process resulted in a subset of 344 human-refined captions that serve as high-quality training and evaluation data for grounded captioning models.
    The primary task involves reviewing and refining the caption: (i) adding references to detected objects that the caption may have overlooked; (ii) removing or correcting incorrect object references or misattributed actions; and, (iii) rewriting sections of the caption that contain factual errors or unclear descriptions.

\begin{figure*}[t]
        \centering
        \includegraphics[width=\textwidth]{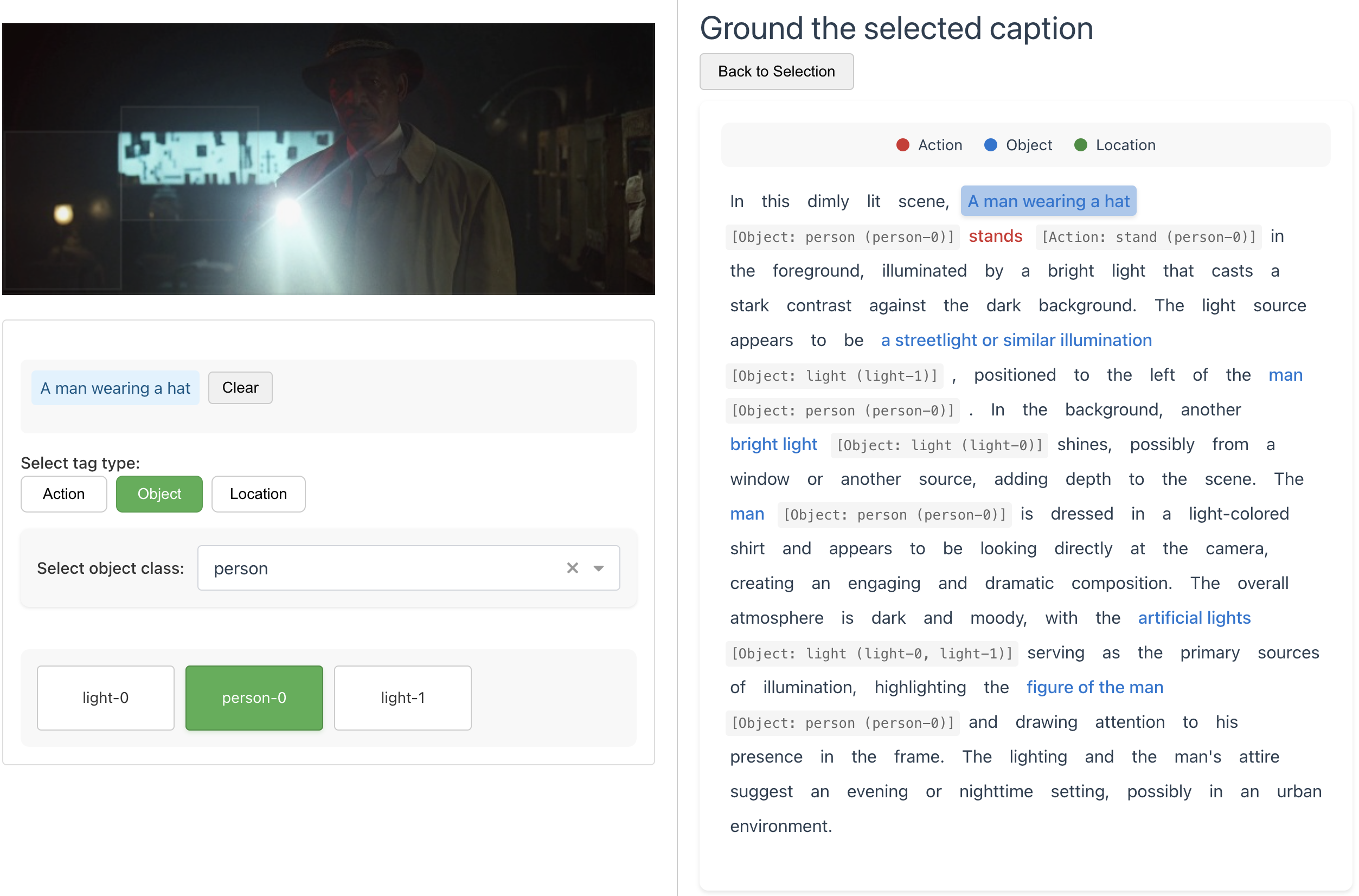}
        \caption{The human annotation interface for caption refinement. The interface displays the image with detected objects (left)
            and the text editor with interactive grounding functionality (right).
            Users can select tag types from a dropdown menu and click on highlighted words to associate them with detected objects, actions or locations.
            Color-coded text spans correspond to the type of grounding tags in the text, blue for objects, red for actiosn and green for locations.}
        \label{fig:annotation-interface}
    \end{figure*}

    The annotation interface (fig.~\ref{fig:annotation-interface}) provides a side-by-side view of the image, caption and object detections.
    Detected objects are highlighted in the image with color-coded bounding boxes that correspond to the grounding tags in the caption text.
    This visual linkage helps annotators quickly identify missing or incorrect references.
    The interface enables direct editing of the caption while maintaining the proper format of grounding tags.

    \section{Grounded METEOR (gMETEOR)}\label{sec:gmeteor}
    Grounded METEOR (gMETEOR) (eq.~\ref{eq:gmeteor}) is a new metric that combines the language quality assessment of METEOR~\cite{banerjee2005meteor} with the grounding accuracy
    measured by F1 score (Section~\ref{subsec:grounded-caption-generation}). %The metric is defined as the harmonic mean of these two components and allows the evaluation of the quality of grounded captions. 
    gMETEOR treats the object detector as an oracle, evaluating captions based on the detected objects rather
    than on the true objects in the image.
    As a consequence, a caption could achieve a perfect score even if it accurately describes incorrectly detected objects, e.g., if the object detector mistakenly identifies a cat as a dog, a caption referring to a dog would be considered correct.

    \begin{equation}
        \text{gMETEOR} = \frac{2 \cdot \text{METEOR} \cdot \text{F1}}{\text{METEOR} + \text{F1}}
        \label{eq:gmeteor}
    \end{equation}

    \section{Baseline Models}\label{sec:baseline-model}

    To establish a baselines, we fine-tuned Pixtral-12B~\cite{pixtral12B}\footnote{Fine-tuned model: \url{https://huggingface.co/daniel3303/PixtralGroundCap}.} and Qwen2.5-VL 7B on GroundCap.
    %We chose Pixtral-12B for its strong performance on vision-language tasks when compared with models from similar size.
    %Pixtral-12B uses a two-component architecture consisting of a vision encoder (a 400-million parameter vision transformer) and a multimodal decoder (a 12-billion parameter 40-layer transformer).
    Our baselines process the full image along with metadata about each detected object (including object ID, class, and normalized coordinates),
    enabling it to generate captions with appropriate grounding tags (\texttt{<gdo>}, \texttt{<gda>}, and \texttt{<gdl>}) that link text spans to specific objects in the image.

    \subsection{Training Methodology}\label{subsec:training-methodology}
    Our model input consists of the full scene image along with metadata for each detected object.
    For each object, we provide its unique identifier, class, and normalized bounding box coordinates [object-id: class, x,y,w,h],
    where (x,y) specifies the top-left corner and (w,h) the width and height of the box.

    We fine-tuned both models using Low-Rank Adaptation (LoRA)~\cite{hu2022lora} with a rank of 16 and alpha scaling factor of 32, targeting both self-attention layers (query, key, value, output projections) and MLP layers (gate, up, down projections) of the language components. This approach reduces memory requirements while maintaining model quality.

    The training follows a two-phase process with different hyperparameters.
    The first phase trains on 52,016 automatic captions using learning rate $2\times 10^{-4}$, batch size 64, and warmup ratio 0.03 for 2 epochs.
    The second phase fine-tunes on 344 human-refined captions using learning rate $2\times 10^{-6}$, batch size 32, with no warmup period to preserve learned features, for 2 epochs.

    We used the AdamW~\cite{adamw2017} optimizer with weight decay 0.01 and bfloat16 precision.
    Training completed in one day using two NVIDIA A100 GPUs (80GB VRAM each).

    \subsection{Evaluation}\label{subsec:evaluation}
    Evaluation of both models was conducted on our test set of 10,000 images,
    comparing performance between the model fine-tuned on automatic captions (Pixtral auto / Qwen auto) and after additional fine-tuning
    on human-refined captions (Pixtral human / Qwen human).

    We include F1 score, precision, recall that measure grounding accuracy,
    BLEU-4~\cite{papineni2002bleu}, METEOR~\cite{banerjee2005meteor}, CIDEr~\cite{vedantam2014cider}, SPICE~\cite{anderson2016spice}
    and ROUGE-L~\cite{lin2004rouge} that evaluate language quality, and gMETEOR that combines grounding accuracy with language quality.

    \begin{table*}[t]
        \centering
        \caption{Evaluation results on the GroundCap test set. Pixtral (auto) and Qwen (auto) refers to the models after fine-tuning on
        automatically generated captions, while Pixtral (human) and Qwen (human) indicates performance after further fine-tuning on human-refined
        captions. We report precision (P), recall (R), F1, BLEU-4 (B-4), METEOR (M), CIDEr (C), SPICE (S), ROUGE-L, and gMETEOR (gM).
        }
        \label{tab:baseline-results}
        \begin{tabular}{lccc|ccccc|c}
            \hline
            \textbf{Model}  & \textbf{P} & \textbf{R} & \textbf{F1} & \textbf{B-4} & \textbf{M} & \textbf{C} & \textbf{S} & \textbf{R} & \textbf{gM} \\
            \hline
            Pixtral (auto)  & 0.61       & 0.95       & 0.70        & 0.18         & 0.24       & 0.46       & 0.30       & 0.37       & 0.35        \\
            Pixtral (human) & 0.58       & 0.96       & 0.69        & 0.19         & 0.23       & 0.51       & 0.30       & 0.37       & 0.35        \\
            Qwen (auto)     & 0.53       & 0.29       & 0.28        & 0.12         & 0.19       & 0.14       & 0.22       & 0.29       & 0.19        \\
            Qwen (human)    & 0.66       & 0.25       & 0.21        & 0.07         & 0.13       & 0.05       & 0.12       & 0.19       & 0.13        \\
            \hline
        \end{tabular}
    \end{table*}

    Our results in Table~\ref{tab:baseline-results} show clear performance differences between the two models.
    Pixtral outperforms Qwen across most metrics, achieving high grounding recall (0.95-0.96) with moderate precision (0.58-0.61),
    resulting in F1 scores around 0.70.
    Qwen shows the opposite pattern: lower recall (0.25-0.29) than precision (0.53-0.66), yielding lower F1 scores (0.21-0.28).
    Pixtral references most detected objects but includes some incorrect groundings, while Qwen is more selective but misses many objects.
    Pixtral's better language quality was expected since the original training captions were generated using Pixtral's policy.
    It achieves BLEU-4 scores of 0.18-0.19 and METEOR scores around 0.24, while Qwen scores substantially lower across all language metrics.
    Pixtral's higher recall does not come from this policy advantage since both models use the same object detections generated by Mask2Former.
    This difference reflects how the models process visual-textual correspondences.

    The comparable performance between automatically trained and human-refined models suggests that our automatic caption generation
    process produces reasonable training data, though there remains room for improvement in both grounding and language generation capabilities.
    These results establish Pixtral as the stronger baseline for GroundCap.

    \section{Human Evaluation}\label{sec:human-evaluation}
    We conducted human evaluation to assess the quality of our grounded captioning system and compare it with baseline approaches.
    This evaluation involved multiple annotators rating different types of captions across several criteria, complemented by an analysis of automatic metrics and an
    exploration of using ChatGPT-4o for automated assessment.
    We focused our human evaluation on Pixtral only due to the costs of conducting human assessment across multiple models 
    and because it yielded the best results in our automatic evaluation.

    \subsection{Evaluation Setup}
\begin{figure*}[t]
        \centering
        \includegraphics[width=\textwidth]{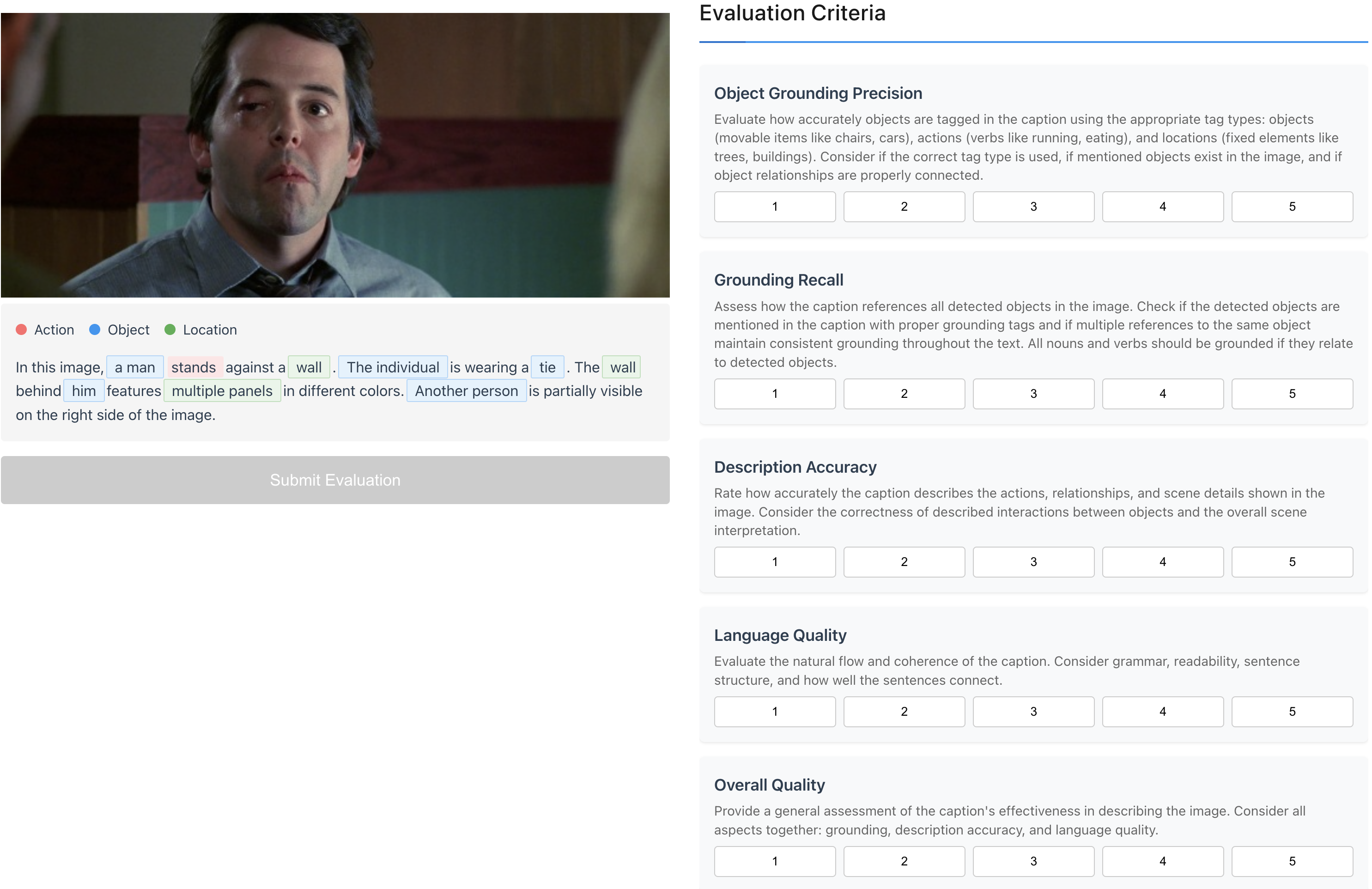}
        \caption{The human evaluation interface shows images with detected objects (top left) and the grounded caption (bottom left). Evaluators rate (right) captions using five criteria on 5-point Likert scales: object grounding precision,
            grounding recall, description accuracy, and language and overall quality.}
        \label{fig:evaluation-interface}
    \end{figure*}

    We compared three caption versions for each image: automatically generated captions from our pipeline, human-refined versions of these captions,
    and outputs from the fine-tuned Pixtral model.
    Each caption was evaluated by three different annotators, resulting in 2,274 total evaluations.
    To prevent bias, captions were presented to evaluators in random order and evaluators could not evaluate captions they had previously annotated.
    Evaluators used an interface showing the image with detected objects alongside the caption being assessed with highlighted grounding tags as shown in fig.~\ref{fig:evaluation-interface}.

    \subsection{Human Assessment}
    Using a 5-point Likert scale (1: Poor, 5: Excellent), evaluators assessed each caption on five criteria.
    Object Grounding Precision evaluates how accurately the caption uses grounding tags (\texttt{<gdo>}, \texttt{<gda>}, \texttt{<gdl>}), including correct tag
    type selection, verification of referenced objects, and semantic correctness of tagged text spans.
    Grounding Recall measures whether all detected objects are appropriately referenced in the caption.
    Description Accuracy focuses on the correctness of described actions, relationships, and scene details,
    ensuring the caption matches the visual content.
    Language Quality considers grammatical correctness, natural flow, and coherence across sentences.
    Finally, Overall Quality provides a holistic assessment combining grounding accuracy, description completeness, and linguistic quality.

    Table~\ref{tab:human-eval-results} presents the human evaluation scores across all criteria.
    Human-refined captions achieved the highest overall quality (4.34), outperforming both automatic captions (4.07) and Pixtral-generated ones (4.22).
    All caption types received high scores for language quality ($>4.8$), demonstrating Pixtral abilities at language generation.
    Human-refined captions showed particular strength in grounding precision (4.23) and recall (4.33), indicating that human refinement improves the accuracy of
    object references while maintaining comprehensive coverage.
    Pixtral-generated captions achieved comparable grounding precision (4.22) to human-refined ones, demonstrating the model's ability to learn
    accurate grounding from the smaller set of human-refined captions.

    \begin{table*}[ht]
        \centering
        \caption{Human ratings for different caption types across evaluation criteria (scale: 1-5)}
        \label{tab:human-eval-results}
        \begin{tabular}{lccccc}
            \hline
            Caption Type    & Object    & Grounding & Description & Language & Overall \\
            & Precision & Recall    & Accuracy    & Quality  & Quality \\
            \hline
            Automatic       & 3.97      & 4.19      & 4.13        & 4.85     & 4.07    \\
            Human-Refined   & 4.23      & 4.33      & 4.17        & 4.82     & 4.34    \\
            Pixtral (human) & 4.22      & 4.19      & 4.08        & 4.91     & 4.22    \\
            \hline
        \end{tabular}
    \end{table*}

    \subsection{Metric Correlation Analysis}
    We analyzed how well traditional automatic metrics align with human judgments of caption quality.
    Table~\ref{tab:correlation-results} shows correlation coefficients between seven metrics and human ratings across our evaluation criteria.
    We include grounding metrics (F1 score, precision, recall) and language quality metrics (BLEU-4, METEOR, ROUGE-L, gMETEOR).
    CIDEr and SPICE were excluded as they require multiple reference captions or large corpora for meaningful computation.

    The analysis revealed consistently weak correlations across all metrics $(\lvert r \rvert < 0.2)$, with the
    strongest correlation being just -0.20 (ROUGE-L with description accuracy).
    Even our proposed gMETEOR metric, which combines grounding and language quality assessment, showed limited correlation with human judgments $(\lvert r \rvert < 0.08)$.
    These results suggest that current automatic metrics fail to capture the aspects of caption quality that humans consider important in grounded captioning,
    highlighting the need for better evaluation methods.

    \begin{table*}[ht]
        \centering
        \caption{Pearson (P) and Spearman (S) correlation coefficients between automatic metrics and human judgments}
        \label{tab:correlation-results}
        \begin{tabular}{lcccccccccc}
            \hline
            & \multicolumn{2}{c}{Object} & \multicolumn{2}{c}{Grounding} & \multicolumn{2}{c}{Description} & \multicolumn{2}{c}{Language} & \multicolumn{2}{c}{Overall} \\
            Metric    & P             & S             & P            & S             & P             & S             & P             & S             & P             & S             \\
            \hline
            F1 Score  & .04           & -.02          & .07          & \textbf{.08}  & -.06          & -.06          & -.07          & .01           & .03           & .03           \\
            Precision & -.02          & -.06          & -.02         & .01           & -.13          & \textbf{-.11} & -.06          & .03           & -.03          & -.02          \\
            Recall    & .03           & \textbf{-.09} & \textbf{.08} & .07           & .06           & -.01          & \textbf{-.09} & \textbf{-.10} & .04           & \textbf{-.05}         \\
            BLEU-4    & -.04          & -.00          & .01          & .04           & -.15          & -.06          & .07           & -.02          & -.09          & .00           \\
            METEOR    & -.05          & .01           & -.02         & .03           & -.20          & -.06          & -.02          & -.05          & -.13          & -.04          \\
            ROUGE-L   & \textbf{-.05} & -.05          & -.01         & .01           & \textbf{-.20} & -.08          & .04           & -.01          & \textbf{-.14} & -.04          \\
            gMETEOR   & -.02          & .00           & .03          & \textbf{.078} & -.15          & -.07          & -.02          & -.02          & -.07          & -.01          \\
            \hline
        \end{tabular}
    \end{table*}

    \subsection{Inter-annotator Agreement}
    To assess the reliability of human evaluations, we measured agreement between annotators using Krippendorff's alpha coefficient.
    Table~\ref{tab:alpha-coefficients} shows the agreement levels across different caption types and evaluation criteria.
    Annotations of Pixtral-generated captions showed the strongest agreement ($\alpha > 0.6$ for all criteria), with particularly high consensus on object identification (0.84),
    description quality (0.86), and overall assessment (0.82).
    This suggests that Pixtral's outputs have consistent, well-defined characteristics that evaluators could assess reliably.

    In contrast, language quality proved the most subjective criterion, with low agreement for automatic captions ($\alpha = 0.03$) and moderate agreement for
    human-refined ones ($\alpha = 0.35$).
    Human-refined captions showed moderate agreement levels ($\alpha \approx 0.4-0.6$) across most criteria, suggesting that while evaluators could identify
    improvements from human refinement, there was less consensus about the degree of improvement.

    \begin{table}[ht]
        \centering
        \caption{Krippendorff's Alpha Coefficients for Inter-Annotator Agreement}
        \label{tab:alpha-coefficients}
        \begin{tabular}{lccccc}
            \hline
            Model           & Object & Grounding & Description & Language & Overall \\
            \hline
            Automatic       & 0.62   & 0.51      & 0.62        & 0.03     & 0.54    \\
            Human-Refined   & 0.48   & 0.47      & 0.64        & 0.35     & 0.41    \\
            Pixtral (human) & 0.84   & 0.80      & 0.86        & 0.60     & 0.82    \\
            \hline
        \end{tabular}
    \end{table}

    \subsection{ChatGPT-4o Evaluation}
    Given the limitations of traditional metrics, we investigated whether large language models could provide more reliable automated evaluation.
    We used ChatGPT-4o~\cite{openai2024chatgpt4o} to evaluate captions using the same criteria and scale as human annotators.
    For each evaluation, we provided ChatGPT-4o with the image, object detections (with coordinates and IDs), the grounded caption, and examples of human evaluations.

    Table~\ref{tab:chatgpt-eval-results} shows that ChatGPT-4o's ratings follow similar patterns to human evaluations, with human-refined captions receiving the highest scores across most criteria.
    Like human evaluators, ChatGPT-4o assigned high language quality scores ($>4.8$) to all caption types, while showing more variation in grounding and description accuracy scores.

    More significantly, ChatGPT-4o's ratings showed strong correlation with human judgments (Table~\ref{tab:lowest-correlations}).
    The strongest correlations appeared in objective criteria: object identification (Pearson=0.81) and description accuracy (Pearson=0.79).
    Even for the more subjective language quality criterion, which showed low inter-annotator agreement, ChatGPT-4o maintained moderate correlation with human ratings (Pearson=0.59).

    These strong correlations, particularly when compared to the poor performance of traditional metrics, suggest that ChatGPT-4o could provide a reliable and scalable approach for evaluating grounded image captions.

    \begin{table*}[ht]
        \centering
        \caption{ChatGPT Ratings for Each Criteria (scale: 1-5)}
        \label{tab:chatgpt-eval-results}
        \begin{tabular}{lccccc}
            \hline
            Caption Type    & Object    & Grounding & Description & Language & Overall \\
            & Precision & Recall    & Accuracy    & Quality  & Quality \\
            \hline
            Automatic       & 4.06      & 4.16      & 4.11        & 4.87     & 4.03    \\
            Human-Refined   & 4.43      & 4.47      & 4.34        & 4.89     & 4.56    \\
            Pixtral (human) & 4.21      & 4.13      & 4.01        & 4.90     & 4.19    \\
            \hline
        \end{tabular}
    \end{table*}

    \begin{table}[ht]
        \centering
        \caption{Correlation between human annotators scores and ChatGPT-4o scores}
        \label{tab:lowest-correlations}
        \begin{tabular}{lccccc}
            \hline
            & Object & Grounding & Description & Language & Overall \\
            \hline
            Pearson  & 0.81   & 0.76      & 0.79        & 0.59     & 0.78    \\
            Spearman & 0.73   & 0.67      & 0.77        & 0.44     & 0.68    \\
            \hline
        \end{tabular}
    \end{table}

    \section{Limitations}\label{sec:limitations}
    GroundCap is based on MovieNet: while it provides a rich source of visual content, it is biased towards movie scenes, and may differ from real-world images in terms of composition, lighting, and camera angles.

    \section{Conclusions and Future Work}\label{sec:conclusions}
    This paper introduced GroundCap, a novel dataset for grounded captioning that provides detailed descriptions
    of visual scenes grounded on detected objects, actions, and locations using an unified grounding framework that
    maintains object identity across multiple references.
    The proposed gMETEOR which combines language quality with grounding accuracy, providing improved quality assessment.
    Additionally, a baseline model for grounded captioning on GroundCap and conducted a human evaluation study
    to assess the quality of our automatically generated, baseline model, and human-refined captions.
    Our baseline model achieves strong grounding recall (0.96) while maintaining good language quality (METEOR score of 0.23).
    Human evaluation demonstrates the effectiveness of our approach, with human-refined captions receiving high ratings across all criteria (4.34/5.0 overall quality).
    Manual inspection revealed that the lowest-performing captions are often frames
    where few or no objects were detected or the image contents were unclear (e.g., in blurry frames).
    We found that ChatGPT-4o evaluations strongly correlated with human judgments (0.78 Pearson correlation for overall quality), suggesting its potential as a scalable evaluation tool, especially given the poor correlation between traditional metrics and human assessment.
    These results establish GroundCap as a valuable resource for advancing research in visually grounded language generation.

    There are several limitations of our work.
    First, GroundCap is derived from MovieNet and consequently inherits biases toward cinematic content (scenes with professional lighting, camera angles,
    and compositions) that may not reflect the diversity of real-world images.
    Second, our object detection system, while effective, occasionally produces false positives or misses objects, which propagates errors to the caption
    generation process.
    Another limitation is that our current approach does not fully address the temporal aspect of visual understanding as the
    captions describe static frames rather than actions unfolding over time.
    This represents an important area for future development, particularly for video understanding applications.

    Future work could explore the extension of GroundCap content from existing image captioning datasets like MS COCO~\cite{lin2014microsoft}
    or Flickr30k~\cite{young2014image}, addressing potential biases toward cinematic content.
    Another direction for future research is extending the ID-based grounding system to handle multiple sequential frames.
    This would enable the development of temporally consistent grounded video descriptions that track objects and their interactions over
    time.
    The development of end-to-end models that can simultaneously detect objects, assign consistent IDs, and generate grounded captions
    without relying on separate object detection systems could also be explored.

    \section*{Acknowledgments}\label{sec:acknowledgments}
    Daniel Oliveira is supported by a scholarship granted by Fundação para a Ciência e Tecnologia (FCT), with reference 2021.06750.BD. Additionally, this work was supported by Portuguese national funds through FCT, with reference UIDB/50021/2020.

    \bibliographystyle{elsarticle-num}
    \bibliography{bibliography}

\end{document}